\documentclass[11pt,letterpaper]{article}
\usepackage{coling2016}
\usepackage{times}
\usepackage{latexsym}
\usepackage[english]{babel}

\usepackage{graphicx} 
\graphicspath{{figures/}} 

\usepackage{amsmath} 
\newcommand{\argmax}{\operatornamewithlimits{argmax}}

\usepackage{subcaption}
\usepackage{booktabs}
\usepackage{multirow}

\title{Sentence Similarity Learning by Lexical Decomposition and Composition}

 \author{Zhiguo Wang \and Haitao Mi \and Abraham Ittycheriah \\
          IBM T.J. Watson Research Center \\ Yorktown Heights, NY, USA \\ {\tt \{zhigwang, hmi, abei\}@us.ibm.com}}
\date{}
\begin{document}

\maketitle
\begin{abstract}
Most conventional sentence similarity methods only focus on similar parts of two input sentences,
and simply ignore the dissimilar parts, 
which usually give us some clues and semantic meanings about the sentences. 
In this work, we propose a model to take into account both the similarities and dissimilarities 
by decomposing and composing lexical semantics over sentences. 
The model represents each word as a vector, 
and calculates a semantic matching vector for each word based on all words in the other sentence. 
Then, each word vector is decomposed into a similar component and a dissimilar component based on the semantic matching vector. 
After this, a two-channel CNN model is employed to capture features 
by composing the similar and dissimilar components. 
Finally, a similarity score is estimated over the composed feature vectors. 
Experimental results 
show that 
our model gets the state-of-the-art performance on the answer sentence selection task, 
and achieves a comparable result on the paraphrase identification task.
\end{abstract}

\section{Introduction}
Sentence similarity is a fundamental metric to measure the degree of likelihood between a pair of sentences. 
It plays an important role for a variety of tasks in both NLP and IR communities. 
For example, in paraphrase identification task, 
sentence similarity is used to determine whether two sentences are paraphrases or not~\cite{yin2015convolutional,he2015multi}. 
For question answering and information retrieval tasks, 
sentence similarities between query-answer pairs are used for assessing the relevance 
and ranking all the candidate answers~\cite{severyn2015learning,wang2015faq}.

However, sentence similarity learning 
has following challenges:
\begin{enumerate}
\item There is a lexical gap between semantically equivalent sentences. 
Take the $E_1$ and $E_2$ in Table~\ref{tab:example} for example, 
they have the similar meaning but with different lexicons. 
\item Semantic similarity should be measured at different levels of granularity (word-level, phrase-level and syntax-level). 
E.g., ``not related'' in $E_2$ is an indivisible phrase when matching with ``irrelevant'' in $E_1$ 
(shown in square brackets). 
\item The dissimilarity (shown in angle brackets) between two sentences is also a significant clue 
\cite{qiu2006paraphrase}. 
For example, by judging the dissimilar parts, we can easily identify 
that $E_3$ and $E_5$ share the similar meaning ``The study is about salmon", because ``sockeye" belongs to the salmon family, and ``flounder" does not. 
Whereas the meaning of $E_4$ is quite different from $E_3$, which emphasizes ``The study is about red (a special kind of) salmon", because both ``sockeye" and ``coho" are in the salmon family.
How we can extract and utilize those information becomes another challenge.
\end{enumerate}

In order to handle the above challenges, researchers have been working on  sentence similarity algorithms for a long time.
To bridge the lexical gap (challenge 1), some word similarity metrics were proposed to 
match different but semantically related words. 
Examples include knowledge-based metrics~\cite{resnik1995using} 
and corpus-based metrics~\cite{jiang1997semantic,yin2015convolutional,he2015multi}. 
To measure sentence similarity from various granularities (challenge 2), 
researchers have explored features extracted from $n$-grams, continuous phrases, 
discontinuous phrases, and parse trees~\cite{yin2015convolutional,he2015multi,heilman2010tree}. 
The third challenge did not get much attention in the past, 
the only related work of \newcite{qiu2006paraphrase} 
explored the dissimilarity between sentences in a pair for paraphrase identification task, 
but they require human annotations in order to train a classifier,
and their performance is still below the state of the art.

\begin{table}[t]
\setlength{\tabcolsep}{0pt}
\hspace{-0.5cm}
\centering
\small
\begin{tabular}{l}
\hline
$E_1$ The research is [irrelevant] to sockeye. \\ 
$E_2$ The study is [not related] to salmon. \\ 
\hline \hline
$E_3$ The research is relevant to salmon. \\ 
$E_4$ The study is relevant to sockeye, $\langle$instead of coho$\rangle$. \\ 
$E_5$ The study is relevant to sockeye, $\langle$rather than flounder$\rangle$. \\ 
\hline
\end{tabular}
\caption{Examples for sentence similarity learning, where sockeye means ``red salmon'', and coho means ``silver salmon''. 
``coho'' and ``sockeye'' are in the salmon family, while ``flounder'' is not.}
\label{tab:example}
\end{table}

In this paper, we propose a novel model to tackle all these challenges jointly 
by decomposing and composing lexical semantics over sentences. 
Given a sentence pair, the model represents each word as a low-dimensional vector (challenge 1), 
and calculates a semantic matching vector for each word based on all words in the other sentence (challenge 2). 
Then based on the semantic matching vector, each word vector is decomposed into two components: 
a similar component and a dissimilar component (challenge 3). 
We use similar components of all the words to represent the similar parts of the sentence pair, 
and dissimilar components of every word to model the dissimilar parts explicitly. 
After this, a two-channel CNN operation is performed to compose the similar and dissimilar components into a feature vector (challenge 2 and 3). 
Finally, the composed feature vector is utilized to predict the sentence similarity. 
Experimental results on two tasks show that 
our model gets the state-of-the-art performance on the answer sentence selection task, 
and achieves a comparable result on the paraphrase identification task.

In following parts, we start with a brief overview of our model (Section~\ref{sec:overview}), 
followed by the details of our end-to-end implementation (Section~\ref{sec:details}).
Then we evaluate our model on answer sentence selection 
and paraphrase identifications tasks (Section~\ref{sec:exps}).

\section{Model Overview}
\label{sec:overview}


In this section, we propose a sentence similarity learning model to tackle all three challenges (mentioned in Section 1). To deal with the first challenge, we represent each word as a distributed
vector, so that we can calculate similarities for formally different but semantically related words. To tackle the second challenge, we assume that each word can be semantically matched by several words in the other sentence, and we calculate a semantic matching vector for each word vector based on all the word vectors in the other side. To cope with the third challenge, we assume that each semantic unit (word) can be partially matched, and can be decomposed into a similar component and a dissimilar component based on its semantic matching vector.

\begin{figure}[tbp]
\begin{center}
\includegraphics[width=0.5\textwidth]{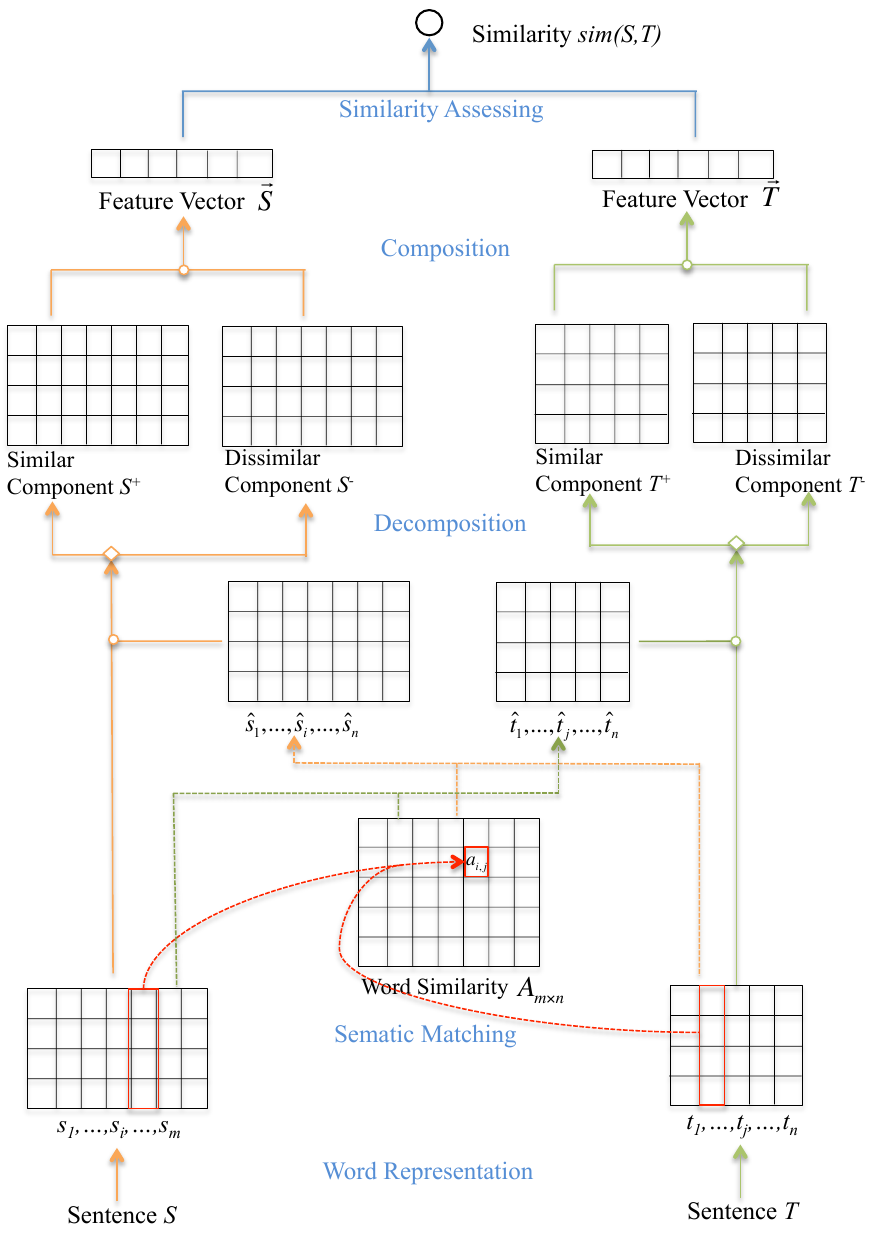}
\end{center}
\caption{Model overview.}
\label{fig:model-overall}
\end{figure}

Figure~\ref{fig:model-overall} shows an overview of our sentence similarity model.
Given a pair of sentences $S$ and $T$, our task is to calculate a similarity score $sim(S,T)$ 
in following steps:

\textbf{Word Representation}. 
Word embedding of \newcite{mikolov2013efficient} is an effective way to 
handle the lexical gap challenge in the sentence similarity task,
as it represents each word with a distributed vector, 
and words appearing in similar contexts tend to have similar meanings~\cite{mikolov2013efficient}. 
With those pre-trained embeddings, we transform $S$ and $T$ into sentence matrixes 
$S=[s_1,...,s_i,...,s_m]$ and $T=[t_1,...,t_j,...,t_n]$, 
where $s_i$ and $t_j$ are \emph{d}-dimension vectors of the corresponding words, 
and $m$ and $n$ are sentence length of $S$ and $T$ respectively.

\textbf{Semantic Matching}. In order to judge the similarity between two sentences, 
we need to check whether each semantic unit in one sentence is covered by the other sentence, or vice versa.
For example, in Table~\ref{tab:example}, to check whether $E_2$ is a paraphrase of $E_1$, 
we need to know the single word ``irrelevant'' in $E_1$ is matched or covered
by the phrase ``not related'' in $E_2$. 
In our model, we treat each word as a primitive semantic unit,  
and calculate a semantic matching vector $\hat{s_i}$ for each word $s_i$ by composing part or full word vectors in the other sentence $T$.
In this way, we can match a word $s_i$ to a word or phrase in $T$. 
Similarly, for the reverse direction, 
we also calculate all semantic matching vectors $\hat{t_j}$ in $T$. 
\begin{equation}
\begin{split}
\hat{s_i} = f_{match}(s_i, T) \hspace{10mm} \forall s_i \in S \\
\hat{t_j} = f_{match}(t_j, S) \hspace{10mm} \forall t_j \in T
\end{split}
\label{equ:match}
\end{equation}
We explore different $f_{match}$ functions later in Section~\ref{sec:details}.

\textbf{Decomposition}. 
After the semantic matching phase, 
we have the semantic matching vectors of $\hat{s_i}$ and $\hat{t_j}$.
We interpret $\hat{s_i}$ (or $\hat{t_j}$) as 
a semantic coverage of word $s_i$ (or $t_j$) by the sentence $T$ (or $S$).
However, it is not necessary that 
all the semantics of $s_i$ (or $t_j$) are fully covered by $\hat{s_i}$ (or $\hat{t_j}$). 
Take the $E_1$ and $E_2$ in Table~\ref{tab:example} for example, 
the word ``sockeye'' in $E_1$ is only partially matched by the word ``salmon'' (the similar part) in $E_2$, 
as the full meaning of ``sockeye'' is ``red salmon'' (the semantic meaning of ``red'' is the dissimilar part). 
Motivated by this phenomenon, 
our model further decomposes word $s_i$ (or $t_j$), based on its semantic matching vector $\hat{s_i}$ (or $\hat{t_j}$), 
into two components: similar component $s_i^+$ (or $t_j^+$) and dissimilar component $s_i^-$ (or $t_j^-$). 
Formally, we define the decomposition function as:
\begin{equation}
\begin{split}
[s_i^+;s_i^-] = f_{decomp}(s_i, \hat{s_i}) \hspace{10mm} \forall s_i \in S \\
[t_j^+;t_j^-] = f_{decomp}(t_j, \hat{t_j}) \hspace{10mm} \forall t_j \in T
\end{split}
\label{equ:decomp}
\end{equation}

\textbf{Composition}. 
Given a similar component matrix $S^+=[s_1^+, ..., s_m^+]$ (or $T^+=[t_1^+,...,t_n^+]$) 
and a dissimilar component matrix $S^-=[s_1^-, ..., s_m^-]$ (or $T^-=[t_1^-,...,t_n^-]$),
our goal in this step is how to utilize those information.
Besides the suggestion from \newcite{qiu2006paraphrase} 
that the significance of the dissimilar parts alone between two sentences has a great effect of their similarity,
we also think that the dissimilar and similar components 
have strong connections.
For example, in Table \ref{tab:example}, 
if we only look at the dissimilar or similar part alone, 
it is hard to judge which one between $E_4$ and $E_5$ is more similar to $E_3$.
We can easily identify that $E_5$ is more similar to $E_3$, when we consider both the similar and dissimilar parts. 
Therefore, 
our model composes the similar component matrix and dissimilar component matrix 
into a feature vector $\vec{S}$ (or $\vec{T}$) with the composition function:
\begin{equation}
\begin{split}
\vec{S} = f_{comp}(S^+, S^-) \\
\vec{T} = f_{comp}(T^+, T^-)
\end{split}
\label{equ:comp}
\end{equation}

\textbf{Similarity assessing}. 
In the final stage, 
we concatenate the two feature vectors ($\vec{S}$ and $\vec{T}$) and predict the final similarity score:
\begin{equation}
sim(S,T) = f_{sim}(\vec{S}, \vec{T})
\label{equ:sim}
\end{equation}

\section{An End-to-End Implementation}
\label{sec:details}
Section~\ref{sec:overview} gives us a glance of our model. 
In this section, we describe details of each phase.

\subsection{Semantic Matching Functions}
\label{subsec:wordmatch}
This subsection describes our specifications for the semantic matching function $f_{match}$ in Eq. (\ref{equ:match}). 
The goal of $f_{match}$ is to generate a semantic matching vector $\hat{s_i}$ for $s_i$ by composing the vectors from $T$. 

For a sentence pair $S$ and $T$, we first calculate a similarity matrix $A_{m\times n}$,
where each element $a_{i,j} \in A_{m\times n}$ computes the cosine similarity 
between words $s_i$ and $t_j$ as
\begin{equation}
a_{i,j}= \frac{s_i^Tt_j}{\|s_i\|\cdot\|t_j\|} \hspace{10mm} \forall s_i \in S, \forall t_j \in T.
\label{equ:cosine}
\end{equation}
Then, 
we define three different semantic matching functions 
over $A_{m \times n}$: 
\begin{equation}
f_{match}(s_i, T)=
\begin{cases}
\frac{\sum_{j=0}^n a_{i,j}t_j} {\sum_{j=0}^n a_{i,j}} &\emph{global}\\
\frac{\sum_{j=k-w}^{k+w} a_{i,j}t_j} {\sum_{j=k-w}^{k+w} a_{i,j}} &\emph{local-w}\\
t_k &\emph{max}\\
\end{cases}
\label{equ:match2}
\end{equation}
where $k=\argmax_j a_{i,j}$.
The idea of the \emph{global} function is to consider all word vectors $t_j$ in $T$. 
A semantic matching vector $\hat{s_i}$ is a weighted sum vector of all words $t_j$ in $T$,
where each weight is the normalized word similarity $a_{i, j}$.
The \emph{max} function moves to the other extreme. 
It generates the semantic matching vector by selecting the most similar word vector $t_k$ from $T$.
The \emph{local-w} function takes a compromise between \emph{global} and \emph{max}, 
where $w$ indicates the size of the window to consider centered at $k$ (the most similar word position).
So the semantic matching vector is a weighted average vector from $t_{k-w}$ to $t_{k+w}$. 


\subsection{Decomposition Functions}
This subsection describes the implementations for the decomposition function $f_{decomp}$ in Eq. (\ref{equ:decomp}). 
The intention of $f_{decomp}$ is to decompose a word vector $s_j$ based on its semantic matching vector $\hat{s_j}$ 
into a similar component $s_i^+$ and a dissimilar component $s_i^-$, 
where $s_i^+$ indicates the semantics of $s_i$ covered by $\hat{s_i}$ 
and $s_i^-$ indicates the uncovered part. 
We implement three types of decomposition function: \emph{rigid}, \emph{linear} and \emph{orthogonal}.

The \emph{rigid} decomposition only adapts to the \emph{max} version of $f_{match}$. 
First, it detects whether there is an exactly matched word in the other sentence, or $s_i$ equal to $\hat{s_i}$. 
If yes, the vector $s_i$ is dispatched to the similar component $s_i^+$, 
and the dissimilar component is assigned with a zero vector \textbf{0}. 
Otherwise, the vector $s_i$ is assigned to the dissimilar component $s_i^-$. 
Eq. ({\ref{equ:rigid}}) gives the formal definition:
\begin{equation}
\begin{split}
[s_i^+ = s_i; \ s_i^- = \textbf{0}] \hspace{10mm} & if \ s_i = \hat{s_i}\\
[s_i^+ = \textbf{0}; \ s_i^- = s_i] \hspace{10mm} & otherwise \\
\end{split}
\label{equ:rigid}
\end{equation}

The motivation for the \emph{linear} decomposition is that the more similar between $s_i$ and $\hat{s_i}$, 
the higher proportion of $s_i$ should be assigned to the similar component.
First, we calculate the cosine similarity $\alpha$ between $s_i$ and $\hat{s_i}$. 
Then, we decompose $s_i$ linearly based on $\alpha$. 
Eq. (\ref{equ:linear}) gives the corresponding definition:

\begin{equation}
\begin{split}
&\alpha = \frac{s_i^T\hat{s_i}}{\|s_i\|\cdot\|\hat{s_i}\|} \\
&s_i^+ = \alpha s_i \\
&s_i^- = (1-\alpha) s_i \\
\end{split}
\label{equ:linear}
\end{equation}

The \emph{orthogonal} decomposition is to decompose a vector in the geometric space. 
Based on the semantic matching vector $\hat{s_i}$, our model decomposes $s_i$ into a parallel component 
and a perpendicular component. 
Then, the parallel component is viewed as the similar component $s_i^+$, 
and perpendicular component is taken as the dissimilar component $s_i^-$. 
Eq. ({\ref{equ:orthogonal}}) gives the concrete definitions.
\begin{equation}
\begin{split}
&s_i^+ = \frac{s_i \cdot \hat{s_i}} {\hat{s_i} \cdot \hat{s_i}} \hat{s_i} \hspace{10mm} parallel \\
&s_i^- = s_i - s_i^+ \hspace{10mm} perpendicular \\
\end{split}
\label{equ:orthogonal}
\end{equation}

\subsection{Composition Functions}
The aim of composition function $f_{comp}$ in Eq. (\ref{equ:comp})
is to extract features from both the similar component matrix and the dissimilar component matrix. 
We also want to acquire similarities and dissimilarities of various granularity during the composition phase. 
Inspired from \newcite{kim2014convolutional}, we utilize a two-channel convolutional neural networks (CNN) 
and design filters based on various order of $n$-grams, e.g., unigram, bigram and trigram.

The CNN model involves two sequential operations: 
\emph{convolution} and \emph{max-pooling}. 
For the \emph{convolution} operation, we define a list of filters \{\textbf{$w_o$}\}. 
The shape of each filter is $d\times h$, where $d$ is the dimension of word vectors and $h$ is the window size. 
Each filter is applied to two patches (a window size $h$ of vectors) from both similar and dissimilar channels, 
and generates a feature. Eq. (\ref{equ:convolution}) expresses this process.

\begin{equation}
c_{o,i} = f(w_o\ast S^+_{[i:i+h]}+w_o\ast S^-_{[i:i+h]}+b_o)
\label{equ:convolution}
\end{equation}
where the operation $A \ast B$ sums up all elements in $B$ with the corresponding weights in $A$, $S^+_{[i:i+h]}$ 
and $S^-_{[i:i+h]}$ indicate the patches from $S^+$ and $S^-$, $b_o$ is a bias term 
and $f$ is a non-linear function (we use \emph{tanh} in this work). 
We apply this filter to all possible patches, and produce a series of features 
$\vec{c_o}=[c_{o,1}, c_{o,2},...,c_{o,O}]$. 
The number of features in $\vec{c_o}$ depends on the shape of the filter $w_o$ and the length of the input sentence. 
To deal with variable feature size, 
we perform a \emph{max-pooling} operation over $\vec{c_o}$ by selecting the maximum value $c_o=max \ \vec{c_o}$. 
Therefore, after these two operations, each filter generates only one feature. 
We define several filters by varying the window size and the initial values. 
Eventually, a vector of features is captured by composing the two component matrixes, 
and the feature dimension is equal to the number of filters.

\subsection{Similarity Assessment Function}
The similarity assessment function $f_{sim}$ in Eq. (\ref{equ:sim}) 
predicts a similarity score by taking two feature vectors as input. 
We employ a linear function to sum up all the features and apply a \emph{sigmoid} function 
to constrain the similarity within the range [0, 1]. 

\subsection{Training}
We train our sentence similariy model by maximizing the likelihood on a training set. 
Each training instance in the training set is represented as a triple ($S_i$, $T_i$, $L_i$), 
where $S_i$ and $T_i$ are a pair of sentences, and $L_i \in \{0,1\}$ indicates the similarity between them. 
We assign $L_i=1$ if $T_i$ is a paraphrase of $S_i$ for the paraphrase identification task, 
or $T_i$ is a correct answer for $S_i$ for the answer sentence selection task. 
Otherwise, we assign $L_i=0$. 
We implement the mathematical expressions with Theano~\cite{Bastien-Theano-2012} 
and use Adam~\cite{kingma2014adam} for optimization.

\section{Experiment}
\label{sec:exps}
\subsection{Experimental Setting}
We evaluate our model on two tasks: answer sentence selection and paraphrase identification. The answer sentence selection task is to rank a list of candidate answers based on their similarities to a question sentence, and the performance is measured by mean average precision (MAP) and mean reciprocal rank (MRR). We experiment on two datasets: \emph{QASent} and \emph{WikiQA}. The statistics of the two datasets can be found in \newcite{yang2015wikiqa}, where \emph{QASent} \cite{wang2007jeopardy} was created from the TREC QA track, and \emph{WikiQA} \cite{yang2015wikiqa} is constructed from real queries of Bing and Wikipedia. The paraphrase identification task is to detect whether two sentences are paraphrases based on the similarity between them. The metrics include the accuracy and the positive class $F_1$ score. We experiment on the Microsoft Research Paraphrase corpus (MSRP) \cite{dolan2004unsupervised}, which includes 2753 true and 1323 false instances in the training set, and 1147 true and 578 false instances in the test set. We build a development set by randomly selecting 100 true and 100 false instances from the training set. In all experiments, we set the size of word vector dimension as $d=$300, and pre-train the vectors with the \emph{word2vec} toolkit \cite{mikolov2013efficient} on the English Gigaword (LDC2011T07).

\subsection{Model Properties}
There are several alternative options in our model, e.g., the semantic matching functions, the decomposition operations, and the filter types. The choice of these options may affect the final performance. In this subsection, we present some experiments to demonstrate the properties of our model, and find a good configuration that we use to evaluate our final model. All the experiments in this subsection were performed on the \emph{QASent} dataset and evaluated on the development set.

\begin{figure}[tbp]
\begin{center}
\includegraphics[width=0.9\textwidth]{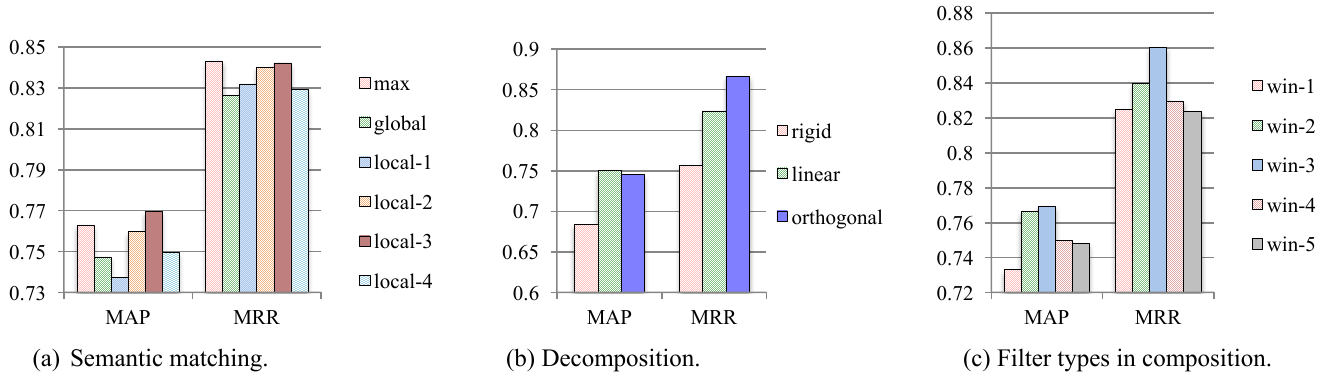}
\end{center}
\caption{Influence of different configuration.}
\label{fig:config-types}
\end{figure}

First, we evaluated the effectiveness of various semantic matching functions. We switched the semantic matching functions among \{\emph{max}, \emph{global}, \emph{local-l}\}, where $l \in$ \{1, 2, 3, 4\}, and fixed the other options as: the \emph{linear} decomposition, the filter types including \{unigram, bigram, trigram\}, and 500 filters for each type. Figure \ref{fig:config-types} (a) presents the results. We found that the \emph{max} function worked better than the \emph{global} function on both MAP and MRR. By increasing the window size, the \emph{local-l} function acquired progressive improvements when the window size is smaller than 4. But after we enlarged the window size to 4, the performance dropped. The \emph{local-3} function worked better than the \emph{max} function in term of the MAP, and also got a comparable MRR. Therefore, we use the \emph{local-3} function in the following experiments.

Second, we studied the effect of various decomposition operations. We varied the decomposition operation among \{\emph{rigid}, \emph{linear}, \emph{orthogonal}\}, and kept the other options unchanged. Figure \ref{fig:config-types} (b) shows the performance. We found that the \emph{rigid} operation got the worst result. This is reasonable, because the \emph{rigid} operation decomposes word vectors by exactly matching words. The \emph{orthogonal} operation got a similar MAP as the \emph{linear} operation, and it worked better in term of MRR. Therefore, we choose the \emph{orthogonal} operation in the following experiments. 

Third, we tested the influence of various filter types. We constructed 5 groups of filters: \emph{win-1} contains only the unigram filters, \emph{win-2} contains both unigram and bigram filters, \emph{win-3} contains all the filters in \emph{win-2} plus trigram filters, \emph{win-4} extends filters in \emph{win-3} with 4-gram filters, and \emph{win-5} adds 5-gram filters into \emph{win-4}. We generate 500 filters for each filter type (with different initial values). Experimental results are shown in Figure \ref{fig:config-types} (c). At the beginning, adding higher-order ngram filters was helpful for the performance. The performance reached to the peak, when we used the \emph{win-3} filters. After that, adding more complex filters decreased the performance. Therefore, the trigram is the best granularity for our model. In the following experiments, we utilize filter types in \emph{win-3}.

\subsection{Comparing with State-of-the-art Models}
In this subsection, we evaluated our model on the test sets of \emph{QASent}, \emph{WikiQA} and \emph{MSRP}.

\textbf{\emph{QASent} dataset.}  Table \ref{tab:qasent} presents the performances of the state-of-the-art systems and our model, where the performances were evaluated with the standard trec\_eval-8.1 script \footnote{http://trec.nist.gov/trec\_eval/}. Given a pair of sentences, \newcite{severyn2015learning} employed a CNN model to compose each sentence into a vector separately, and joined  the two sentence vectors to compute the sentence similarity. Because only the sentence-level granularity was used, the performance is much lower (the second row of Table \ref{tab:qasent}). After adding some word overlap features between the two sentences, the performance was improved significantly (the third row of Table \ref{tab:qasent}). Therefore, the lower-level granularity is an indispensable factor for a good performance. \newcite{wang2015faq} conducted word alignment for a sentence pair based on word vectors, and measured the sentence similarity based on a couple of word alignment features. They got a slightly better performance (the fourth row of Table \ref{tab:qasent}), which indicates that the vector representation for words is helpful to bridging the lexical gap problem. \newcite{cicero2016attentive} introduced the attention
mechanism into the CNN model, and learnt sentence representation by considering the influence of the other sentence. They got better performance than all the other previous work. Our model makes use of all these useful factors and also considers the dissimilarities of a sentence pair. We can see that our model (the last row of Table \ref{tab:qasent}) got the best MAP among all previous work, and a comparable MRR than \newcite{cicero2016attentive}.

\textbf{\emph{WikiQA} dataset.} Table \ref{tab:wikiqa} presents the results of our model and several state-of-the-art models. \newcite{yang2015wikiqa} constructed the dataset and reimplemented several baseline models. The best performance (shown at the second row of Table \ref{tab:wikiqa}) was acquired by a bigram CNN model combining with the word overlap features. \newcite{miao2015neural} models the sentence similarity by enriching LSTMs with a latent stochastic attention mechanism. The corresponding performance is given at the fourth row of Table \ref{tab:wikiqa}. \newcite{yin2015abcnn} introduced the attention mechanism into the CNN model, and captured the best performance (the fifth row of Table \ref{tab:wikiqa}). The semantic matching phase in our model is similar to the attention mechanism. But different from the previous models, our model utilizes both the similarity and dissimilarity simultaneously. The last row of Table \ref{tab:wikiqa} shows that our model is more effective than the other models.

\begin{table}
\small
\begin{minipage}{0.45\textwidth}
\centering
\begin{tabular}{l|cc}
    \toprule
    Models & MAP    & MRR    \\
    \midrule
    \begin{tabular}[c]{@{}l@{}}\newcite{severyn2015learning} \\ (CNN only)\end{tabular}              & 0.6709 & 0.7280 \\
    \midrule
    \begin{tabular}[c]{@{}l@{}}\newcite{severyn2015learning}\\  (CNN + sparse features)\end{tabular} & 0.7459 & 0.8078 \\
    \midrule
    \begin{tabular}[c]{@{}l@{}}\newcite{wang2015faq}\\  (Word embedding alignment)\end{tabular} & 0.7460 & 0.8200 \\
    \midrule
    \begin{tabular}[c]{@{}l@{}}\newcite{cicero2016attentive}\\  (Attention-based CNN)\end{tabular} & 0.7530 & \textbf{0.8511} \\
    \midrule
    \midrule
    This work & \textbf{0.7714} & 0.8447 \\
    \bottomrule
\end{tabular}
\caption{Results on the QASent dataset.}
\label{tab:qasent}
\end{minipage}%
\hfill
\begin{minipage}{0.5\textwidth}
\centering
\begin{tabular}{l|cc}
\toprule
Models            & MAP    & MRR    \\
    \midrule
    \begin{tabular}[c]{@{}l@{}}\newcite{yang2015wikiqa} \\ (2-gram CNN)\end{tabular} & 0.6520 & 0.6652 \\
    \midrule
    \begin{tabular}[c]{@{}l@{}}\newcite{cicero2016attentive}\\  (Attention-based CNN)\end{tabular} & 0.6886 & 0.6957 \\
    \midrule
    \begin{tabular}[c]{@{}l@{}}\newcite{miao2015neural} \\ (Attention-based LSTM)\end{tabular} & 0.6886 & 0.7069 \\
    \midrule
    \begin{tabular}[c]{@{}l@{}}\newcite{yin2015abcnn} \\ (Attention-based CNN)\end{tabular} &  0.6921 & 0.7108 \\
    \midrule
    \midrule
This work & \textbf{0.7058} & \textbf{0.7226} \\
    \bottomrule
\end{tabular}
\caption{Results on the WikiQA dataset.}
\label{tab:wikiqa}
\end{minipage}
\end{table}

\textbf{\emph{MSRP} dataset.} Table \ref{tab:paraphrase} summarized the results from our model and several state-of-the-art models. \newcite{yin2015convolutional} employed a CNN model to learn sentence representations on multiple level of granularity and modeled interaction features at each level for a pair of sentences. They obtained their best performance by pretraining the model on a language modeling task (the 3rd row of Table  \ref{tab:paraphrase}). However, their model heavily depends on the pretraining strategy. Without pretraining, they got a much worse performance (the second row of Table  \ref{tab:paraphrase}). \newcite{he2015multi} proposed a similar model to \newcite{yin2015convolutional}. Similarly, they also used a CNN model to extract features at multiple levels of granularity. Differently, they utilized some extra annotated resources, e.g., embeddings from part-of-speech (POS) tags and PARAGRAM vectors trained from the Paraphrase Database \cite{ganitkevitch2013ppdb}. Their model outperformed \newcite{yin2015convolutional} without the need of pretraining (the sixth row of Table \ref{tab:paraphrase}). However, the performance was reduced after removing the extra resources (the fourth and fifth rows of Table \ref{tab:paraphrase}). \newcite{yin2015abcnn} applied their attention-based CNN model on this dataset. By adding a couple of sparse features and using a layerwise training strategy, they got a pretty good performance. Comparing to these neural network based models, our model obtained a comparable performance (the last row of Table \ref{tab:paraphrase}) without using any sparse features, extra annotated resources and specific training strategies. However, the best performance so far on this dataset is obtained by \newcite{ji2013discriminative}. In their model, they just utilized several hand-crafted features in a Support Vector Machine (SVM) model. Therefore, the deep learning methods still have a long way to go for this task.

\begin{table}[tbp]
\centering
\begin{tabular}{l|cc}
\toprule
Models & Acc  & F1    \\
\midrule
\newcite{yin2015convolutional} (without pretraining) & 72.5 & 81.4  \\
\midrule
\newcite{yin2015convolutional} (with pretraining) & 78.4 & 84.6  \\
\midrule
\newcite{he2015multi} (without POS embeddings) & 77.8 & N/A   \\
\midrule
\newcite{he2015multi} (without Para. embeddings) & 77.3 & N/A   \\
\midrule
\newcite{he2015multi} (POS and Para. embeddings) & 78.6 & 84.7 \\
\midrule
\newcite{yin2015abcnn} (with sparse features) & 78.9 & 84.8  \\
\midrule
\newcite{ji2013discriminative} & \textbf{80.4} & \textbf{86.0}  \\
\midrule
\midrule
This work & 78.4 & 84.7 \\
\bottomrule
\end{tabular}
\caption{Experimental results for paraphrase identification on MSRP corpus.}
\label{tab:paraphrase}
\end{table}
\section{Related Work}
The semantic matching functions in subsection~\ref{subsec:wordmatch} are inspired from the attention-based neural machine translation \cite{bahdanau2014neural,luong2015effective}. However, most of the previous work using the attention mechanism in only LSTM models. Whereas our model introduces the attention mechanism into the CNN model. A similar work is the attention-based CNN model proposed by \newcite{yin2015abcnn}. They first build an attention matrix for a sentence pair, and then directly take the attention matrix as a new channel of the CNN model. Differently, our model uses the attention matrix (or similarity matrix) to decompose the original sentence matrix into a similar component matrix and a dissimilar component matrix, and then feeds these two matrixes into a two-channel CNN model. The model can then focus much on the interactions between similar and dissimilar parts of a sentence pair.

\section{Conclusion}
In this work, we proposed a model to assess sentence similarity by decomposing and composing lexical semantics. To bridge the lexical gap problem, our model represents each word with its context vector. To extract features from both the similarity and dissimilarity of a sentence pair, we designed several methods to decompose the word vector into a similar component and a dissimilar component. To extract features at multiple levels of granularity, we employed a two-channel CNN model and equipped it with multiple types of ngram filters. Experimental results show that our model is quite effective on both the answer sentence selection task and the paraphrase identification task .
\section*{Acknowledgments}
We thank the anonymous reviewers for useful comments.

\bibliographystyle{acl2016}
\bibliography{qa}

\end{document}